\documentclass[final]{cvpr}

\usepackage{times}
\usepackage{epsfig}
\usepackage{graphicx}
\usepackage{amsmath}
\usepackage{amssymb}
\usepackage{multirow}

\usepackage[pagebackref=true,breaklinks=true,colorlinks,bookmarks=false]{hyperref}
\def\NEW#1{\textcolor{black}{#1}}  

\def\cvprPaperID{11676} 
\def\confYear{CVPR 2021}

\begin{document}
\title{Densely connected multidilated convolutional networks for dense prediction tasks}

\author{Naoya Takahashi, Yuki Mitsufuji\\
Sony Corporation, Japan\\
}

\maketitle

\begin{abstract}
Tasks that involve high-resolution dense prediction require a modeling of both local and global patterns in a large input field. 
Although the local and global structures often depend on each other and their simultaneous modeling is important, many convolutional neural network (CNN)-based approaches interchange representations in different resolutions only a few times. 
In this paper, we claim the importance of a dense simultaneous modeling of multiresolution representation and propose a novel CNN architecture called densely connected multidilated DenseNet (D3Net).
D3Net involves a novel multidilated convolution that has different dilation factors in a single layer to model different resolutions simultaneously. 
By combining the multidilated convolution with the DenseNet architecture, D3Net incorporates multiresolution learning with an exponentially growing receptive field in almost all layers, while avoiding the aliasing problem that occurs when we naively incorporate the dilated convolution in DenseNet.
Experiments on the image semantic segmentation task using Cityscapes and the audio source separation task using MUSDB18 show that the proposed method has superior performance over state-of-the-art methods.


\end{abstract}

\begin{figure*}[t]
  \centering
  \includegraphics[width=\linewidth]{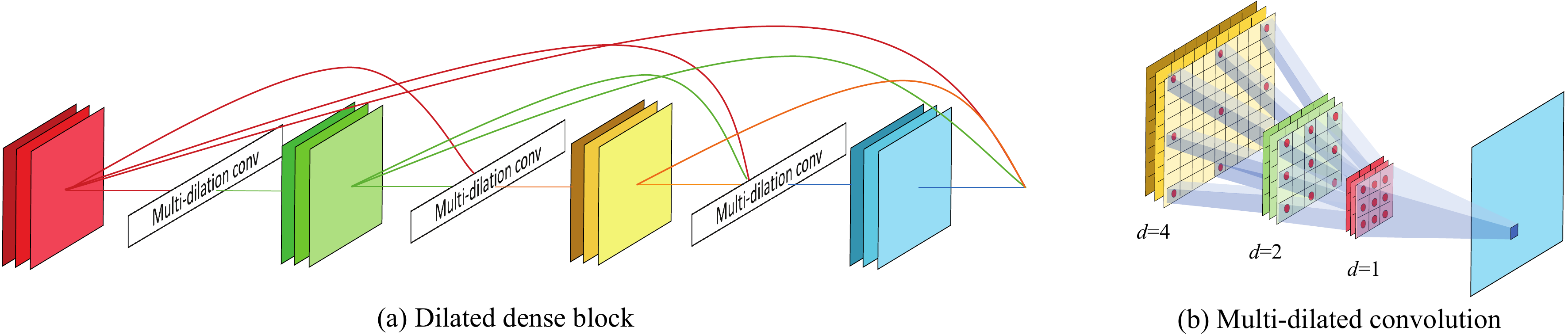}
  \caption{Illustration of D2 block. (a) The connectivity pattern is the same as that in DenseNet except that the D2 block involves the multidilated convolution. (b) Illustration of the multidilated convolution at the third layer. The production of a single feature map involves multiple dilation factors depending on the input channel. For clarity, we omit the normalization and nonlinearity from the illustration.
}
  \label{fig:d2block}
\end{figure*}

\section{Introduction}
\label{sec:intro}
Dense prediction tasks such as semantic segmentation and audio source separation typically accept high-dimensional input data and produce predictions with the same (or similar) dimensions.
To efficiently handle high-dimensional data and model the context that lies in a large field, various neural network architectures have been proposed \cite{Long15,SunXLW19,WangSCJDZLMTWLX19,Takahashi17}.
In particular, convolutional neural networks (CNNs) have become an essential component, and a variety of advanced CNN architectures have been proposed to improve performance on the basis of motivations such as making the networks deeper while improving a gradient flow \cite{He16ResNet,Huang17Densenet,Larsson17}, multibranch convolution \cite{Szegedy15, Szegedy16} and explicitly modeling interchannel dependences of convolutional features \cite{Hu18SEnet}. One key component of these architectures is a \textit{skip connection} that creates short paths from early layers to later layers. In \cite{Huang17Densenet}, a simple yet powerful skip connectivity pattern that connects all preceding layers, called DenseNet, is proposed. Such dense connectivity allows maximum information flow, making CNNs deeper while keeping the model size small by efficiently reusing intermediate representations of preceding layers.

One of the benefits of a deeper CNN is its larger receptive field that allows a large context to be modeled, which is important for tasks that require the utilization of a wide-area or long-term dependence in a high-resolution input. For example, sufficiently large parts of objects have to be modeled for semantic segmentation \cite{Chen2015,Yu16dilation,Chen17Deeplab,zhao2017pspnet,zhao2018psanet,Seif18,WangSCJDZLMTWLX19,fu2018dual,Yuan20OCR, Zhong20SAN}, whereas modeling a long-term dependence is shown to be important for various audio tasks such as audio event recognition and source separation \cite{Takahashi2016, Takahashi2017AENet, Takahashi18MMDenseLSTM}. 
Although the receptive field grows linearly with respect to the number of layers stacked, the simple stacking of convolution layers is not the optimal way to increase it, as too many layers are required to cover a sufficiently large input, which makes the network training difficult.
A popular approach to incorporate a large context with a reasonable model size is to repeatedly downsample intermediate network outputs and apply operations in lower resolution representations. 
In dense prediction tasks, the low-resolution representations are again upsampled to recover the resolution lost while carrying over the global perspective from downsampled layers \cite{Noh15, Ronneberger15UNet,Long15, WangSCJDZLMTWLX19,Takahashi18MMDenseLSTM}. 
Another approach is dilated convolution, where dilation factors are set to grow exponentially as layers are stacked; and therefore, the networks cover a large receptive field with a small number of layers. Dilated convolution is shown to be effective for many tasks that require high-resolution dense predictions \cite{Yu16dilation,Chen2017DeepLab,Aaron2016WN}. 
Most previously proposed CNN architectures interchange information in different resolutions only a few times, e.g., once or a few times at the end of the network \cite{Long15,zhao2017pspnet,zhao2018psanet}, or once at the beginning or end of each module \cite{Ronneberger15UNet,WangSCJDZLMTWLX19}. 
However, since the local and global patterns can depend on each other, i.e., a local structure can be more accurately estimated by knowing a global structure and vice versa, a more frequent (dense) interchange of information among representations in multiple resolutions could be beneficial.

In this work, we propose a novel CNN architecture for densely incorporating representations in multiple resolutions.
We combine advantages of the dense skip connections and dilated convolution, and propose a novel network architecture called the multidilated dense block (D2 block). To appropriately combine them, we propose a multidilated convolution layer that has multiple dilation factors within a single layer. A dilation factor depends on which skip connection the feature maps come from, as shown in Fig.\ref{fig:d2block}. Multidilated convolution can prevent the occurrence of aliasing that occurs when a standard dilated convolution is applied to feature maps whose receptive field is smaller than the dilation factor. 
Furthermore, we propose a nested architecture of multidilated dense blocks to effectively repeat dilation factors multiple times with dense connections that ensure sufficient depth, which is required for modeling each resolution. 
We call the nested architecture densely connected multidilated DenseNet (D3Net)
\footnote{Code is available at \url{https://github.com/sony/ai-research-code/tree/master/d3net}}.

Although neural network architecture search (NAS) has been actively investigated to automatically find a suitable network architecture \cite{Liu_2019_CVPR,Nekrasov_2019_CVPR}, it is often difficult to identify the key element for achieving good performance from the learnt architecture.  We believe that this work provides another insight into the design of CNN architectures for dense prediction tasks, namely, the frequent interchange of information in multiple resolutions.
The contributions of this work are as follows:
(i) We claim the importance of the dense multiresolution representation learning and propose the D2 block that combines dense skip connections with dilated convolution. The D2 block incorporates a novel multidilated convolution that enables multiresolution information interchange in most of the layers while avoiding the aliasing problem that occurs in a naive way of incorporating dilation in DenseNet. 
(ii) We further introduce a nested architecture of multidilated dense blocks called the D3 block to effectively apply different dilation factors multiple times to provide a sufficient modeling capacity in each resolution. 
(iii) We conduct intensive experiments on two dense prediction tasks in different domains (image semantic segmentation and audio source separation) and show the effectiveness of the proposed methods. The proposed architecture exhibits superior performance over state-of-the-art baselines in both tasks, demonstrating its generality against the task type and data domain.

\begin{figure*}[t]
  \centering
  \includegraphics[width=\linewidth]{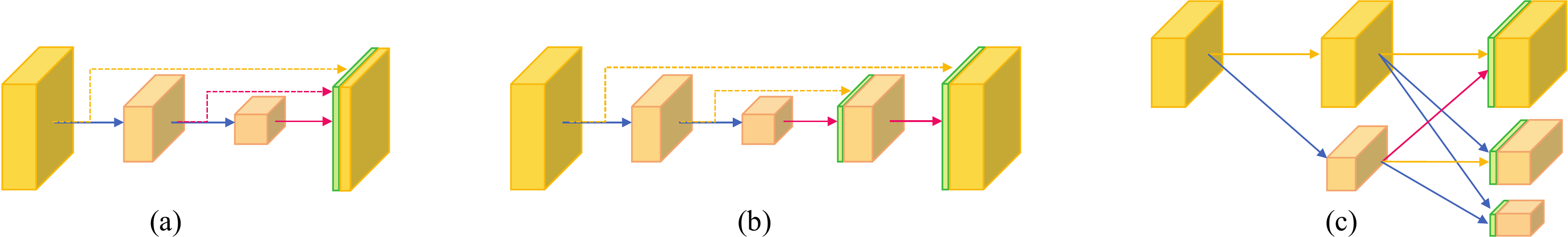}
  \caption{Strategies for multiscale representation integration. The yellow box indicates a composition of convolution layers, which operates in a single resolution. The green box depicts a layer that integrates feature maps from different resolutions. (a) Feature maps in multiple scales are integrated at the end \cite{Long15}. (b) Feature maps in the lower scale are sequentially recover a higher scale by integrating the feature maps from the higher scale in the early layer \cite{Ronneberger15UNet,Newell2016Hourglass}. (c) Features in different resolutions are first processed in parallel and integrated at the end of each stage \cite{SunXLW19,WangSCJDZLMTWLX19}.}
  \label{fig:multires}
\end{figure*}
\section{Related works}
The motivation of our work is to combine the advantages of dense skip connectivity and dilated convolution to enable multiresolution modeling with an exponentially growing receptive field while appropriately avoiding the aliasing problem. Here, we review related works on these aspects. 
\vspace{3mm}\\
\textbf{Dense skip connection} \hspace{1mm}
Dense skip connections from early layers promote the reuse of feature maps, efficient parameter usage, and gradient information flow. DenseNet has the most dense connectivity pattern (i.e., all layers with same feature-map size are connected to each other) and shows excellent performance in image classification tasks \cite{Huang17Densenet}. Larsson \etal proposed another simple connectivity pattern called FractalNet, in which layers are connected in fractal manner \cite{Larsson17}. Dual path networks combine DenseNet and ResNet to enjoy the advantage of the dense connectivity with the concatenation of feature maps and residual blocks, which involve the addition of feature maps \cite{Chen2017DualPN}. 
\vspace{3mm}\\
\textbf{Large receptive field} \hspace{1mm}
The importance of a large receptive field was addressed in many tasks that involve high-dimensional data including image super-resolution \cite{Seif18LRF}, semantic segmentation \cite{Chen17Deeplab,zhao2017pspnet,zhao2018psanet,Chen2015,Yu16dilation}, and audio source separation \cite{Takahashi18MMDenseLSTM}. The theoretical receptive field size of CNNs does not directly represent the context size that CNNs use. Zhou \etal showed that the empirical receptive field of CNNs is much smaller than the theoretical one, especially in deeper layers \cite{Zhou2015}. 
Therefore, network architectures that efficiently incorporate context information in a large field attract great interest and many approaches have been proposed including the incorporation of the dilated convolution \cite{Yu16dilation,Chen2017DeepLab}, the aggregating of downsized feature maps \cite{zhao2017pspnet}, and the use of the attention mechanism \cite{zhao2018psanet,fu2018dual,Zhong20SAN,zhang2020resnest}.
\vspace{3mm}\\
\textbf{Dilated convolution and aliasing} \hspace{1mm}
Aliasing is a well-known effect in signal processing, in which the signal over the Nyquist frequency becomes indistinguishable with lower frequency after (sub-)sampling. The aliasing causes artifacts such as the Moir\'{e} pattern in the image domain or audible noise in the audio domain. Therefore, a low-pass filter for anti-aliasing is typically applied before sampling to remove the signal with a frequency higher than the Nyquist frequency. The effect of pooling-based subsampling in CNN-based speech recognition was studied and a performance drop caused by aliasing was observed \cite{Gong18}. The dilated convolution involves the subsampling of input feature maps and can cause aliasing \NEW{\cite{Wang18HDC}}. To avoid this problem, most CNN architectures that involve dilated convolution are carefully designed to allow earlier layers to learn appropriate anti-aliasing filter if necessary, i.e., standard convolutions are applied before dilated convolutions with fixed dilation factor \cite{Chen17Deeplab,Yang2020DilatedInception,Wang18HDC,Li20PSConv}, or the dilation factors is gradually increased as the layer goes deeper \cite{Yu16dilation, Aaron2016WN}. 
A naive combination of DenseNet with dilation has already been proposed \cite{Fuchs19}, where dilated convolutions are used and the dilation factor was set to one at the initial layer and doubled as the layer goes deeper. However, this approach has significant aliasing due to skip connections, as discussed in Sec. \ref{sec:d2}. 
\vspace{3mm}\\
\textbf{Multiresolution modeling} \hspace{1mm} 
Fusing local and global information is important especially for dense prediction tasks, since both local and global structures have to be recovered. In the fully convolutional network (FCN) \cite{Long15}, feature maps in different resolutions from early layers are aggregated at the end of the network (Fig.~\ref{fig:multires}(a)). Another common strategy used in, for instance, UNet \cite{Ronneberger15UNet} and Hourglass\cite{Newell2016Hourglass}, is the sequential upsampling of feature maps while combining the feature maps from early downsampling paths with skip connection, as shown in Fig.~\ref{fig:multires}(b), which aggregates multiresolution information at few concatenation points. HRNet \cite{SunXLW19,WangSCJDZLMTWLX19} involves another strategy for the aggregation of feature maps (Fig. \ref{fig:multires}(c)). It is composed of several stages: in each stage, feature maps in different resolutions are first processed by CNNs individually and then aggregated by matching the resolution with other resolutions with up- or downsampling at the end of each stage.  In these approaches, feature maps in different resolutions are fused only a few times. 
Another stage-wise aggregation was proposed in \cite{Yu18}, where the feature maps in different resolutions are aggregated iteratively and hierarchically.
In contrast, our method fuses feature maps with multiple resolutions in almost all layers (except the first layer of D2 blocks and few other layers such as $1\times1$ convolution layers).
Multibranch convolution can also be considered as multiresolution modeling when the convolution in each branch operates in a different resolution. In \cite{Shi2017DilatedInception,Chen2017DeepLab,Yang2020DilatedInception,Schuster19}, dilated convolutions with different dilation factors are applied in parallel to the same feature maps and combined in a multibranch convolution module called the dilated inception (DI) or SDC layer. The set of dilation factors is the same for all modules. 
In \cite{Li20PSConv}, Poly-Scale convolution (PSConv) arranges multiple dilation factors periodically along with input channels.
In contrast, the dilation factors in multidilated convolution depends on the skip connection as shown in Fig. \ref{fig:d2block}, and their range grows exponentially as the layer goes deeper.
\NEW{Moreover, DI, SDC, and PSConv themselves cannot solve the aliasing problem and they require several layers before these modules to circumvent it. Therefore, they cannot fuse very local information in the first few layers, and multiresolution modeling can be performed only on the feature maps that are anti-aliased by the first several layers that possibly remove high-frequency components. In contrast, proposed D3 block can be directly applied to the input, which enables to fuse very local and global information.
}
MSDenseNet \cite{Huang18MSDense} also involves a frequent two resolution fusion. However, the architecture is not suitable for dense prediction tasks as there is no information flow from low- to high-resolution feature maps.  In Res2Net \cite{Gao21Res2Net}, multi-scale feature maps in a single resolution are aggregated.

\begin{figure}[t]
  \centering
  \includegraphics[width=\linewidth]{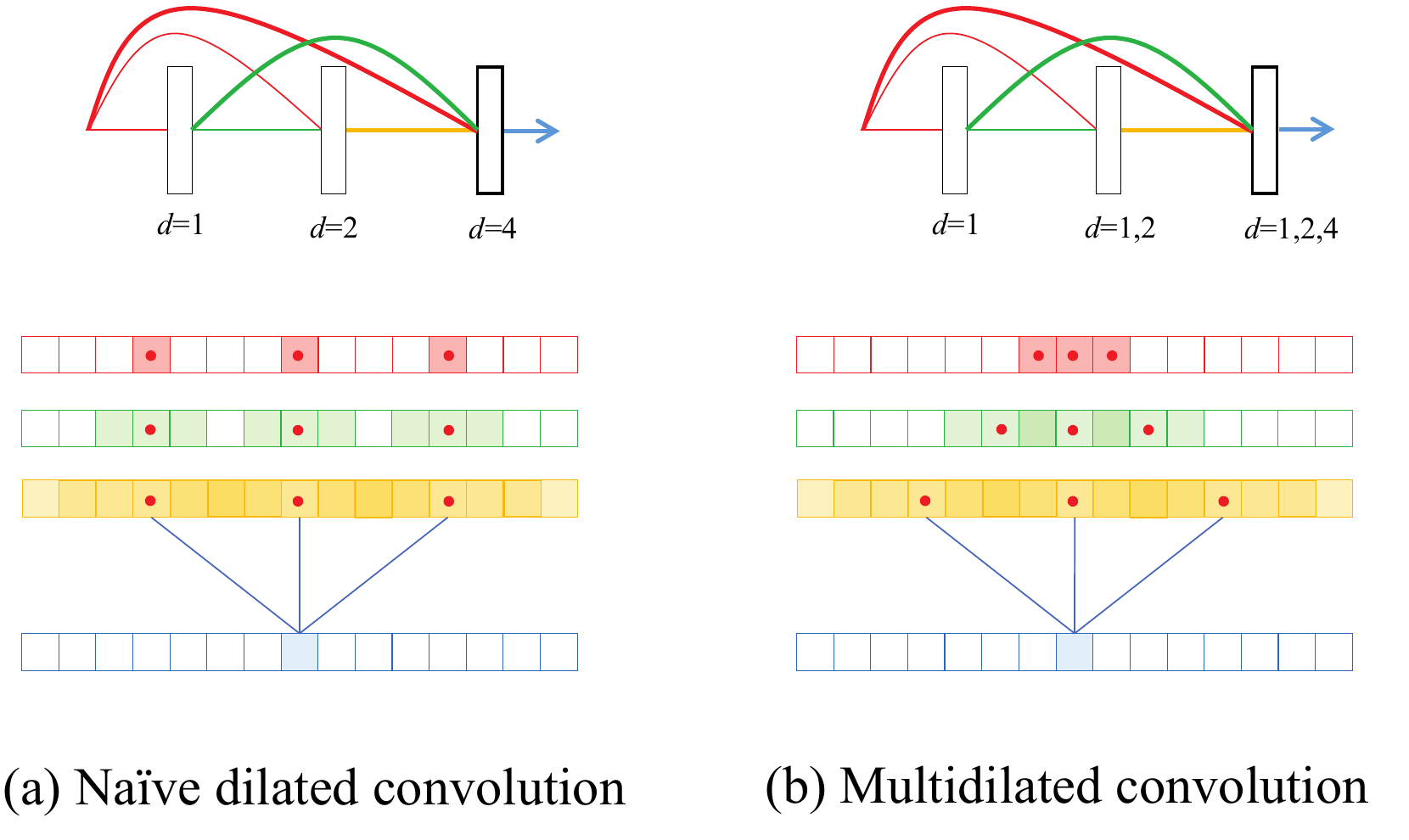}
  \caption{Visualization of receptive fields at the third layer of (a) naive integration of dilated convolution and (b) proposed multidilated convolution (in the case of one dimension). The filter size is 3. Red dots denote the points on which filters are applied, and the colored background shows the receptive field covered by the red dots. In (a), convolution kernels for skip connections have \textit{blind spots} in their receptive fields, while the multidilated convolution (b) appropriately changes the dilation factor to avoid them.}
  \label{fig:d2rf}
\end{figure}

\section{Multidilated convolution for DenseNet}
\label{sec:d2}
In DenseNet, the outputs of the $l$th convolutional layer $x_l$ are computed using $3\times3$ convolution filters $w_l$ and outputs of all preceding layers as
\begin{equation}
x_l = \psi([x_0, x_1, \cdots, x_{l-1}]) \circledast w_l,
\end{equation}
where $\psi()$ denotes the composite operation of batch normalization and ReLU nonlinearity, $[x_0, x_1, \cdots, x_{l-1}]$ the concatenation of feature maps from $0, \cdots, l-1$ layers ($x_0$ is the input), and $\circledast$ the convolution. $x_{l>0}$ has $k$ feature maps and $k$ is the \textit{growth rate}.
A naive way of incorporating dilated convolution is to replace the convolution $\circledast$ with the dilated convolution $\circledast_d$ with the dilation factor $d=2^{l-1}$. However, this causes a severe aliasing problem; for instance, at the third layer, input is subsampled at four sample intervals without any anti-aliasing filtering because of the skip connections. Only the path that passes through all convolution operations without any skip connection covers the input field without omission, and all other paths from skip connections have \textit{blind spots} in their receptive fields that inherently make it impossible for appropriate ant-aliasing filters to be learned in the preceding layers (Fig.~\ref{fig:d2rf}(a)). 
To overcome this problem, we propose the multidilated convolution $\circledast^m_l$ defined as
\begin{equation}
Y_l \circledast^m_l k_l = \sum_{i=0}^{l-1} y_i \circledast_{d_i} w^i_l,
\end{equation}
where $Y_l = [y_0, \cdots, y_{l-1}] = \psi([x_0, \cdots, x_{l-1}])$ is the composite layer output,  $w^i_l$ the subset of filters that corresponds to the $i$th skip connection, and $d_i = 2^i$. As depicted in Fig.~\ref{fig:d2rf}(b), DenseNet with the proposed multidilated convolution has different dilation factors depending on which layer the channel comes from. This allows the receptive field to cover the input field without the loss of coverage between the samples to which the filters are to be applied and, hence, to learn proper filters to prevent aliasing. 

One advantage of the multidilated convolution is its capability to integrate information from the very local to global information of an exponentially large receptive field within a single layer. Combined with the dense skip connection topology, D2 blocks can perform multiresolution modeling in all layers (except the first layer). 
This fast information flow with dense skip connections and the dense (frequent) information interchange among representations in a wide range of resolutions provide a more flexible capability of modeling a relationship between local and global structures.

Note that the multidilation convolution is not equivalent to the multibranch convolution, where convolutions with different dilation factors are applied to the same input feature maps, similar to the Inception block \cite{Szegedy15, Szegedy16}, as it again causes the aliasing problem when combined with the dense skip connection topology.
\begin{figure}[t]
  \centering
  \includegraphics[width=\linewidth]{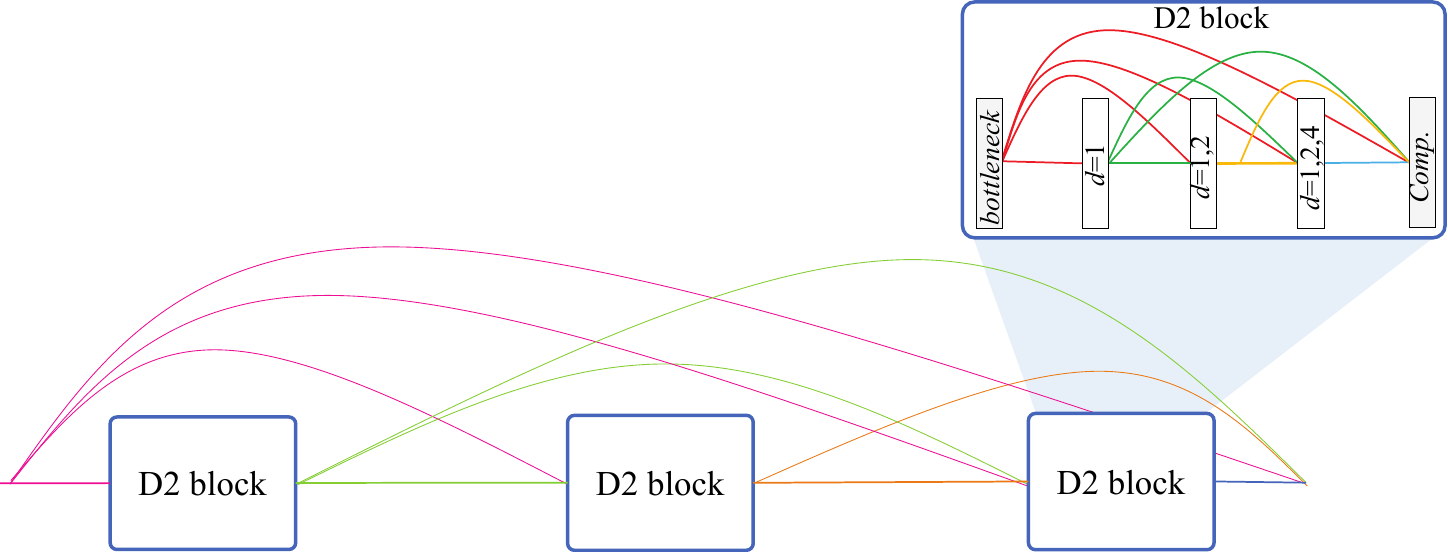}
  \caption{D3 block densely connects D2 blocks with repeated dilation patterns.}
  \label{fig:d3block}
\end{figure}

\section{D3Net}
\label{sec:d3}
Although the D2 block provides an exponentially large receptive field as the number of layers increases, it is also worthwhile to provide sufficient flexibility to transform feature maps in each resolution. In WaveNet \cite{Aaron2016WN}, dilation factors are reset to one after several layers are stacked and repeated; that is, the dilation factor in the $l$th layer is given by $d_l=2^{l-1~mod~M}$, where $mod$ is the modulo operation and $M$ is the number of layers at which the dilation factor is doubled. Inspired by this work, we propose a nested architecture of D2 blocks, as shown in Fig.~\ref{fig:d3block}. D2 blocks are considered single composite layers and are densely connected in the same way as within the D2 block itself. With the $M$ D2 blocks nested, the multidilated convolution operates at each resolution at least $M$ times, providing a flexible modeling capability at each resolution. We refer to this nested architecture as the D3 block. 

Inspired by the DenseNet-BC architecture \cite{Huang17Densenet}, we also employ two channel-reduction mechanisms to mitigate the excessive increase in the number of channels and thus improve computational efficiency. First, we adopt bottleneck layers that reduce the number of input channels using  $1\times1$ convolution at the beginning of each D2 block. In our experiment, bottleneck layers were set to produce $4k$ feature maps, where $k$ is the growth rate, and such layers are placed only when the input channel to the D2 block is greater than $4k$. Second, we compress the output channels at the end of each D2 block by $1\times1$ convolution to produce $cm$ channels, where $0<c<1$ is the compression rate and $m$ is the number of channels before the compression. Alternatively, we can simply pass the outputs of the last $N$ layers to the next D2 block. 
In our experiment, we used the former approach for semantic segmentation and the latter approach for audio source separation.
Note that without the channel reduction layer, the D3 architecture is reduced to standard dense connections with repeated multidilation factors.

\section{Implementation details}
Our proposed D3 block can be integrated with CNN architectures commonly used in image classification (e.g., VGG \cite{Zisserman2015VGG}, ResNet \cite{He16ResNet}), image segmentation (e.g., FCN \cite{Long15} and deconvolution-based approaches \cite{Noh15,Ronneberger15UNet, Fu19}), and audio tasks \cite{Takahashi17} by replacing the series of convolution layers in the same resolution with a D3 block. We call a CNN architecture that uses D3 blocks as D3Net. When D3Net involves downsampling between D3 blocks, we adopt a transition layer which is composed of a $1\times1$ convolution layer followed by $2\times2$ average pooling. In the transition layer, the number of output channels is compressed to half of the input channels, as performed in DenseNet \cite{Huang17Densenet}.
In summary, a D3 block is characterized with a set of parameters $(M,L,k,B,c)$, where $M$ denotes the number of D2 blocks in a D3 block (Fig. \ref{fig:d3block}), $L$ the number of layers in each D2 block, $k$ the growth rate, $B$ the number of bottleneck layer channels (which is set to $4k$ in our experiments), and $c$ the compression rate.

\section{Experiments}
We evaluate the proposed method on two dense prediction tasks, namely, image semantic segmentation and audio source separation, to show the generality of the proposed approach against the task and data domain.

\begin{figure}[t]
  \centering
  \includegraphics[width=\linewidth]{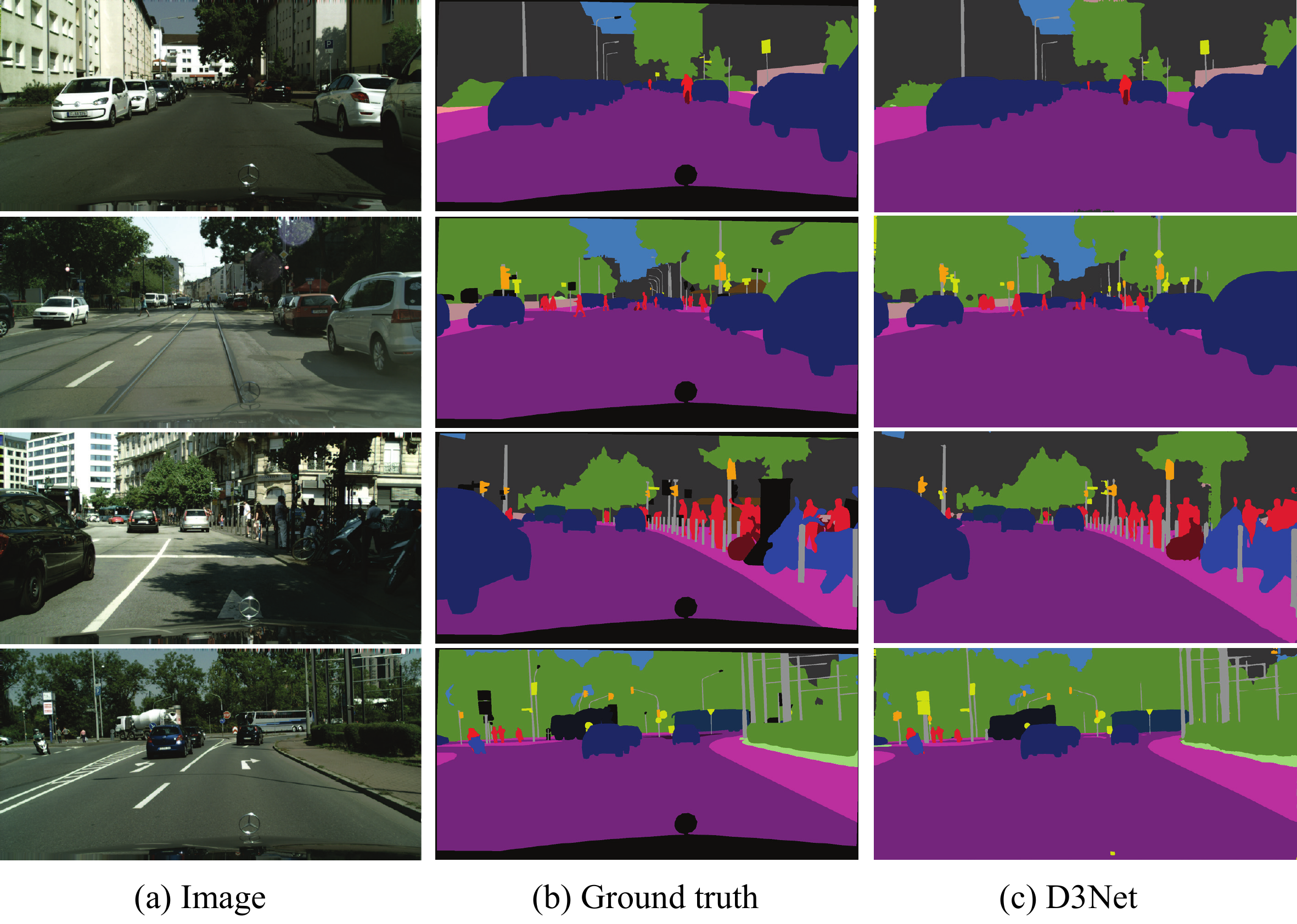}
  \caption{Qualitative examples of Cityscapes results on \textit{val} set.}
  \label{fig:results_cityscapes}
\end{figure}
\subsection{Semantic segmentation}
The goal of semantic segmentation is to assign a class label to each pixel, as shown in Fig.~\ref{fig:results_cityscapes}. Since our contribution is the CNN architecture, we mainly focus on the evaluation of backbone networks. 
To this end, unless otherwise noted, all experiments including baselines are conducted under the same training/testing setup using the MMSegmentation\footnote{https://github.com/open-mmlab/mmsegmentation} framework.
\vspace{3mm}\\
\textbf{Dataset.} \hspace{1mm}
We use the Cityscapes dataset \cite{Cordts2016Cityscapes}, which contains 5,000 images collected from street scenes in 50 different cities with high quality pixel-level annotation.
The images are divided into 2,975, 500, 1,525 for training, validation, and testing, respectively. We did not use coarsely annotated images.
Following the evaluation protocol in \cite{Cordts2016Cityscapes}, 19 categories are used for evaluation and we report the mean of class-wise intersection over union (mIoU).
\vspace{3mm}\\
\textbf{Model architecture} \hspace{1mm}
D3Net consists of two $3\times3$ convolution layers followed by four D3 blocks with transition layers in between. Here, we refer to the downsample ratio as ``scale''; therefore D3 blocks operate in four different scales. Outputs of D3 blocks in each scale are combined and passed to a decode head in the same way as in \cite{Long15,WangSCJDZLMTWLX19}, i.e., feature extraction layers formed by $1\times1$ convolution are applied to the outputs of each D3 block to collect features from all scales, and the features in a lower scale are rescaled by bilinear upsampling to match the highest scale. Finally, another $1\times1$ convolution is performed on the concatenation of the rescaled features to mix the information in four representations.

We consider two D3Nets. The smaller architecture, denoted as D3Net-S, employs D3 blocks of $(M,L,k,c) = (4,8,36,0.2)$, while the larger architecture, D3Net-L, uses D3 blocks of $(M,L,k,c) = (4,10,64,0.2)$. The number of feature maps extracted from each scale using the feature extraction layers are $(32,40,64,128)$ for D3Net-S, and $(32,48,96,192)$ for D3Net-L. 
\vspace{3mm}\\
\textbf{Training} \hspace{1mm}
We follow the same training protocol as in \cite{zhao2017pspnet,zhao2018psanet}. The data augmentation of random horizontal flip, random cropping (from $1024\times2048$ to $512\times1024$), and random scaling in the range of $[0.5, 2]$ are performed.
The stochastic gradient descent with a momentum of 0.9 and a
weight decay of 0.0005 is used for optimization. 
The ``poly'' learning rate policy with a base learning rate of 0.01 and a power of 0.9 is used for dropping the learning rate. All the models are trained on the \textit{training} set with a batch size of 8 and the synchronized batch normalization \cite{Zhang18SyncBN}.
\vspace{3mm}\\
\textbf{Ablation study} \hspace{1mm}
First, we focus on the evaluation of the proposed multidilated convolution with dense connections (D2 block) and the nested architecture (D3 block). To this end, we consider four baselines. To highlight the effect of the multidilated convolution, we consider models with the same architecture as D3Net-S but replace the multidilated convolution with a standard convolution (without dilation) and a standard dilated convolution, whose dilation factors $d$ are equal to the maximum dilation factor in the corresponding multidilated convolution layer in D3Net, e.g., $d=(1,2,4,8,1,2,4,\cdots)$. For the evaluation of the nested architecture, we consider a model that replaces the D3 block with a standard dense block (with BC layers) \cite{Huang17Densenet}. For a fair comparison, we design the dense block to have a similar parameter size to D3Net-S by either keeping the growth rate and fitting the number of layers, or keeping the number of layers nearly the same and fitting the growth rate. This results in two DenseNet baselines, DenseNet-133 that has 16 layers for each Dense block with the growth rate of 36, and DenseNet-189 that has 23 layers for each dense block with the growth rate of 23 (the number after DenseNet- indicates the total number of layers). For reference, we also evaluate commonly used backbone networks. All networks are trained from scratch for 40,000 iterations. Table \ref{tab:cityscapesAbration} shows the mIoU scores on the validation set.
\begin{table}[t]
\caption{\label{tab:cityscapesAbration}Ablation study on Cityscapes \textit{val} set. D-ResNet stands for Dilated-ResNet}
\small
\centering
\begin{tabular}{ c | c | c } 
\hline
Backbone &	\#param. &	mIoU  \\
\hline
D-ResNet-50 \cite{He16ResNet} &	49.5M &	59.7\\
D-ResNet-101  \cite{He16ResNet} &		68.5M &	62.4\\
HRNetV2-W18 \cite{WangSCJDZLMTWLX19} &	9.6M &	62.7\\
HRNetV2-W48 \cite{WangSCJDZLMTWLX19} &	65.9M &	67.7 \\
\hline
D3Net-S without dilation  &	9.7M &	62.3 \\
D3Net-S standard dilation  &	9.7M &	57.9 \\
DenseNet-133 ($k=36$)	&	10.2M & 57.6\\
DenseNet-189 ($k=23$) &	10.0M  &	54.9\\
\hline
D3Net-S &		9.7M  &	65.1\\
D3Net-L &	38.7M  &	\textbf{68.1}\\
\hline
\end{tabular}
\end{table}

\begin{table}[t]
\caption{\label{tab:cityscapesVal}Cityscapes \textit{val} set results. No test-time augmentation (multiscale, flipping) is applied. $\dagger$ denotes results reported in reference papers.}
\resizebox{\linewidth}{!}{
\begin{tabular}{ c | c c | c } 
\hline
Method &    Backbone &	\#param. &	mIoU  \\
\hline
DeepLabV3 \cite{Chen17Deeplab}&	D-ResNet-50 &	 68.1M &	79.3	 \\
DeepLabV3 \cite{Chen17Deeplab}&	D-ResNet-101 & 87.1M &	80.2	 \\
DeepLabV3 \cite{Chen17Deeplab} &	ResNeSt-101 \cite{zhang2020resnest} & 90.8M & 79.7	 \\
DeepLabV3+$\dagger$ \cite{Chen18DeepLab3plus}  & Xception-71 & 43.5M & 79.6\\
PSPNet \cite{zhao2017pspnet} &	D-ResNet-101 &	68.0M &	79.8	 \\
PSANet \cite{zhao2018psanet}&	D-ResNet-101 & 78.1M &	79.3	 \\
Auto-DeepLab-L$\dagger$ \cite{Liu_2019_CVPR} & -   &  44.4M & 80.3 \\
FCN &	D-ResNet-50 &	49.5M &	73.6 \\
FCN &	D-ResNet-101 &	68.5M &	75.1 \\
FCN &	HRNetV2-W18 \cite{WangSCJDZLMTWLX19}&	9.6M &	78.7 \\
FCN &	HRNetV2-W48 \cite{WangSCJDZLMTWLX19}&	65.9M &	79.9 \\
OCRNet &	HRNetV2-W48 \cite{WangSCJDZLMTWLX19}&	70.3M &	80.7 \\
\hline
FCN &	D3Net-S &	9.7M &	79.5 \\
FCN &	D3Net-L &	38.7M & 80.6 \\
OCRNet &	D3Net-L &	42.3M & \textbf{81.2} \\
\hline
\end{tabular}
}
\end{table}

\begin{table}[t]
\centering
\caption{\label{tab:cityscapesTest}Results on Cityscapes \textit{test} set. Baseline results are from original papers. All models are trained on the \textit{train} set without using coarse data.}
\small
\begin{tabular}{ c | c | c  } 
\hline
 & Backbone &	mIoU \\
\hline
PSPNet \cite{zhao2017pspnet}&	D-ResNet-101 &	78.4 \\
PSANet \cite{zhao2018psanet}&	D-ResNet-101 &	78.6 \\
PAN \cite{Li18PAN} &	D-ResNet-101 &	78.6 \\
AAF \cite{Ke18AAF}&	D-ResNet-101 &	79.1 \\
HRNetV2 \cite{WangSCJDZLMTWLX19} &	HRNetV2-W48 &	80.4 \\
\hline
D3Net (FCN) &	D3Net-L &	\textbf{80.8} \\
\hline
\end{tabular}
\end{table}

D3Net-S (with the proposed multidilated convolution) performs significantly better than D3Net-S without dilation and D3Net-S with the standard dilation, improving mIoU by 2.8 points. This highlights the effectiveness of the multidilated convolution in dense connections. Interestingly, D3Net-S with the standard dilation performs significantly worse than the model without dilation. This is probably due to the aliasing problem since a large dilation factor is applied directly to the initial feature map, as discussed in Sec.~\ref{sec:d2}. 
D3Net-S without dilation outperforms DenseNet-133 by 4.7 points, where both models have the same growth rate and no dilation. This could be because the receptive field of DenseNet-133 covers the entire input only in the last few layers, which did not provide a sufficient capacity to model global information. On the other hand, D3Net-S without dilation still largely outperforms DensNet-189, which has almost the same number of layers as D3Net-S.  This is probably due to followings: the growth rate in DenseNet-189 had to be a smaller to have the similar parameter size and the receptive field of DenseNet-189 is still smaller than D3Net-S without dilation as DenseNet involves more $1\times1$ convolutions, which does not increase the receptive field. These results highlight the efficiency of the proposed nested architecture, the D3 block.

D3Nets-L exhibits the best performance among all baselines with a much smaller number of parameters than current state-of-the-art backbone networks, such as HRNetV2-W48. D3Net-S outperforms dilated ResNet101 with nearly a seven times smaller parameter size, showing the parameter efficiency of the proposed architecture.
\vspace{3mm}\\
\textbf{Comparison with state-of-the-art approaches} \hspace{1mm}
Next, we compare D3Net with state-of-the-art approaches in Table \ref{tab:cityscapesVal}. Again, our focus is on the evaluation of D3Net as a backbone, and we train all models in the same setup (expect methods denoted with $\dagger$) to eliminate the effect of hyperparameter difference that mainly comes from computational resources such as the batch size.  We initialize all backbone networks with weights pretrained on ImageNet \cite{ILSVRC15} and trained 80K iterations.
Among backbone networks in the FCN approach, D3Net-L shows superior performance over all baselines with a much smaller number of parameters than HRNetV2p-W48\cite{WangSCJDZLMTWLX19}, D-ResNet101, or D-ResNet50. By combining with the object-contextual representation (OCR)
scheme \cite{Yuan20OCR}, D3Net further improves the performance, obtaining the best result of 81.2\% in this experiment.
In Table \ref{tab:cityscapesTest}, we also show the results for the \textit{test} set. All results are with six scales and flipping. For this experiment, we train D3Net-L for 160K iterations with a batch size of 12. All other settings are the same as those in previous experiments. 
Baseline results are from the original papers. The proposed method again outperforms all baselines that were trained on the \textit{train} set.



\subsection{Audio source separation}
\label{sec:exSS}
\begin{figure}[t]
  \centering
  \includegraphics[width=\linewidth]{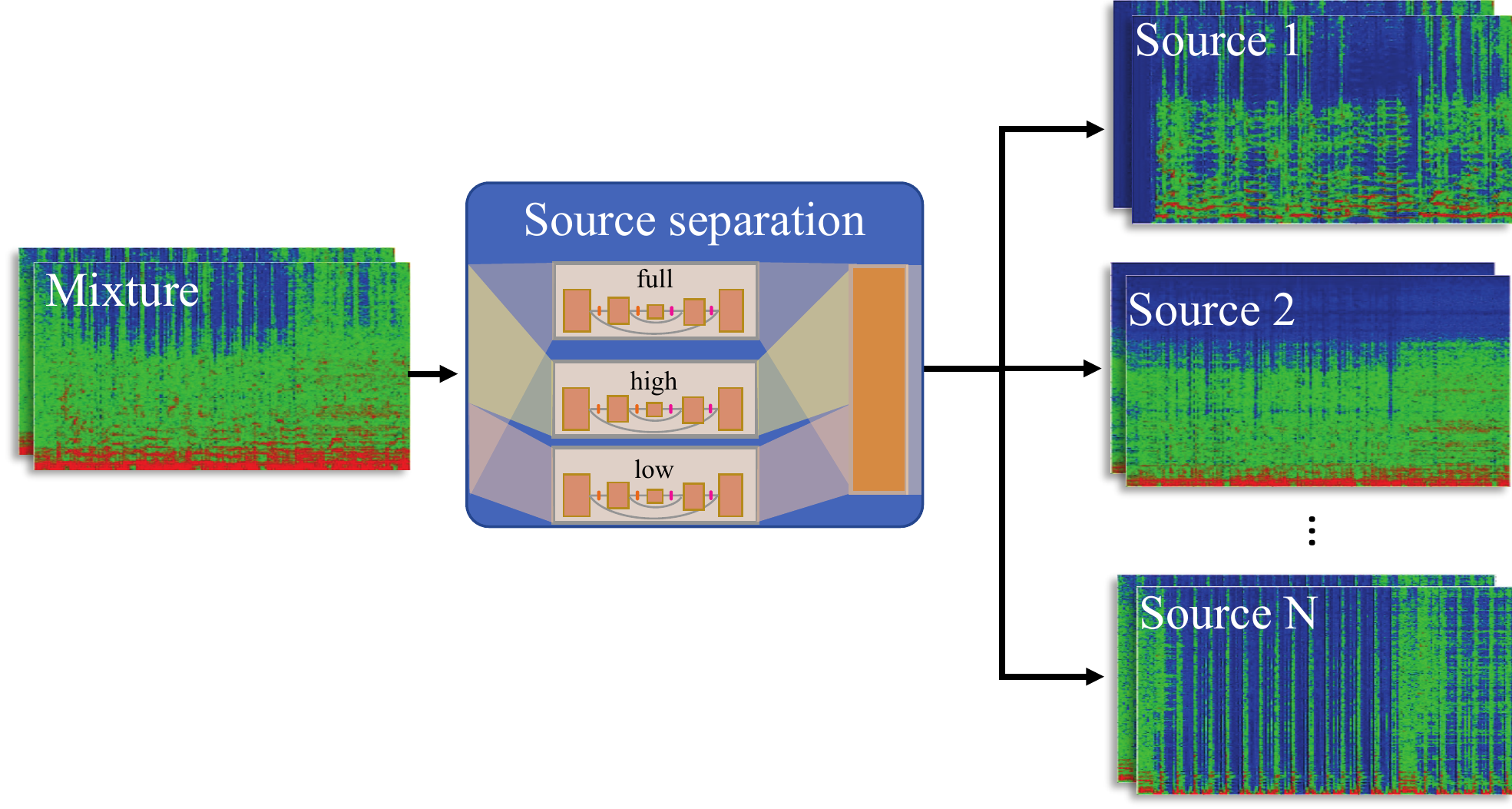}
  \caption{Illustration of audio source separation in STFT domain.}
  \label{fig:sourceseparation}
\end{figure}

To show the generality of the proposed method in a different domain, we conduct experiments on an audio source separation task, where the goal is to separate source signals from their mixture. Recently, CNN-based methods have been intensively studied \cite{Nachmani20,Liu19,stoter19,Samuel20,defossez2019demucs,Takahashi18MMDenseLSTM}. In most methods, a time domain signal is transformed by short-time Fourier transform (STFT) and source separation is performed in the magnitude STFT domain. In this case, the audio source separation problem is similar to an image segmentation problem, i.e., a model accepts two-dimensional magnitude STFT maps and predicts the source magnitude for each time-frequency bin (cf. pixels in an image), as shown in Fig. \ref{fig:sourceseparation}. However, there are three major differences. First, source separation is a regression problem rather than a pixel-wise classification problem, as the model is trained to estimate the source magnitude STFT.  Second, when multiple sources are in the same time-frequency bin, they are summed in a complex STFT domain, unlike objects in an image, where a front object can hide an object at the back (occlusion). 
Since only magnitude is considered in complex STFT, the mixing behavior becomes more complex. Third, in the STFT domain, the translation invariant property is not globally satisfied along with the frequency axis, although local translation along with frequency and translation along the time axis are invariant. 
\vspace{3mm}\\
\textbf{Dataset} \hspace{1mm}
We use the MUSDB18 dataset prepared for the SiSEC 2018 challenge \cite{sisec2018}. 
In this dataset, approximately 10 hours of professionally recorded 150 songs in the stereo format at 44.1kHz are available. For each song, a mixture and its four sources, {\it bass, drums, other}, and {\it vocals}, are provided; thus, the task is to separate the four sources from the mixture. We adopted the official split of 100 and 50 songs for the {\it Dev} and {\it Test} sets, respectively.
STFT magnitude frames of the mixture, windowed at 4096 samples with 75\% overlap, with data augmentation \cite{Uhlich17} are used as inputs. 
\vspace{3mm}\\
\textbf{Training} \hspace{1mm}
The four networks for each source instrument are trained to estimate the source spectrogram by minimizing the mean square error with the Adam optimizer for 50 epochs. The patch length is set to 256 frames; thus, the dimensions of input were $2\times256\times2049$. The batch size is set to 6. The learning rate is initially set to 0.001 and annealed to 0.0001 at 40 epochs.
\vspace{3mm}\\
\textbf{Model architecture} \hspace{1mm}
Following \cite{Takahashi17,Takahashi18MMDenseLSTM}, in which the best results obtained in SiSEC 2018 were reported, we use the multiscale multiband architecture in which band-dedicated modules and a full band module, each with a bottleneck encoder--decoder architecture with skip connections, are placed. The network configuration is shown in Table \ref{tab:d3arch}. The network outputs are used to calculate the multichannel Wiener filter (MWF) to obtain the final separations, as commonly performed in frequency domain audio source separation methods \cite{Takahashi18MMDenseLSTM, Uhlich17, Liu19, Samuel20}.
\begin{table}[t]
\caption{\label{tab:mss} SDRs for MUSDB18 dataset. 
'*' denotes the method operating in the time domain.}
\vspace{2mm}
\centering{
\resizebox{\linewidth}{!}{
\begin{tabular}{ c | c c c c c c} 
\hline
\multicolumn{1}{c|}{} & \multicolumn{6}{c}{SDR in dB}\\
Method      &	Vocals	&	Drums	& Bass & Other & Acco. & Avg.\\
\hline\hline
TAK1 {\scriptsize (MMDenseLSTM)} \cite{Takahashi18MMDenseLSTM}\ 	&	6.60 & 6.43 & 5.16 & 4.15 & 12.83 &5.59\\
UHL2 {\scriptsize(BLSTM ensemble)} \cite{Uhlich17}\ 	&	5.93 & 5.92 & 5.03 & 4.19 & 12.23 &5.27\\
GRU dilation 1 \cite{Liu19}\ 	& 6.85 & 5.86 & 4.86 & \textbf{4.65} & 13.40 & 5.56\\
UMX \cite{stoter19}\ 	& 6.32 & 5.73 & 5.23 & 4.02 & - & 5.33\\
demucs* \cite{defossez2019demucs}\ 	& 6.29 & 6.08 & 5.83 & 4.12 & - & 5.58\\
Meta-TasNet* \cite{Samuel20} 	& 6.40 & 5.91 & 5.58 & 4.19 & - & 5.52\\
Nachmani \etal* \cite{Nachmani20} & 6.92 & 6.15 &  \textbf{5.88} & 4.32 & - & 5.82\\
\hline
D3Net without dilation      & 6.86 & 6.37	& 4.97 & 4.21 & 13.19 & 5.60\\
D3Net standard dilation	& 7.12	& 6.61 & 5.19 & 4.53  & 13.39 & 5.86\\
{\bf D3Net} (proposed)	& \textbf{7.24}	& {\bf 7.01} & 5.25 & 4.53 & {\bf 13.52} & \textbf{6.01}\\
\hline
\end{tabular}
}
}
\end{table}
\begin{table*}[t]
    \caption{\label{tab:d3arch} Proposed architectures. All D3 blocks have 3$\times$3 kernels with growth rate $k$, $L$ layers, and $M$ D2 blocks.}
    \vspace{2mm}
    \centerline{
      \footnotesize
      \tabcolsep=3px
      \begin{tabular}{ c | c | c | c | c | c | c | c | c | c | c} 
        \hline
        \multirow{2}{*}{Layer} & \multirow{2}{*}{scale} & \multicolumn{3}{c|}{Vocals, Other} & \multicolumn{3}{c|}{Drums} & \multicolumn{3}{c}{Bass} \\
        \cline{3-11}
        & & low & high & full & low & high & full & low & high & full \\
        \hline
        band split index & \multirow{3}{*}{1} & 1-256 & 257-1600 &  - & 1-128 & 128-1600 &  - & 1-192 & 192-1600 &  -\\
        conv (t$\times$f,ch) &  & 3$\times$3, 32 & 3$\times$3, 8  & 3$\times$3, 32 & 3$\times$3, 32 & 3$\times$3, 8  & 3$\times$3, 32 & 3$\times$3, 32 & 3$\times$3, 8  & 3$\times$3, 32 \\
        D3 block 1 (k,L,M) & & 16, 5, 2 & 2, 1, 1 & 13, 4, 2 & 16, 5, 2 & 2, 1, 1 & 13, 4, 2 & 16, 5, 2 & 2, 1, 1 & 10, 4, 2\\
        \hline                 
        down sample  & \multirow{2}{*}{$\frac{1}{2}$} &  \multicolumn{3}{c|}{avg. pool $2\times2$} & \multicolumn{3}{c|}{avg. pool $2\times2$} & \multicolumn{3}{c}{avg. pool $2\times2$}\\
        \cline{3-11}
        D3 block 2 (k,L,M) & & 18, 5, 2 & 2, 1, 1 & 14, 5, 2  & 18, 5, 2 & 2, 1, 1 & 14, 5, 2 & 18, 5, 2 & 2, 1, 1 & 10, 5, 2\\
        \hline
        down sample  & \multirow{2}{*}{$\frac{1}{4}$} &  \multicolumn{3}{c|}{avg. pool $2\times2$} & \multicolumn{3}{c|}{avg. pool $2\times2$} & \multicolumn{3}{c}{avg. pool $2\times2$}\\
        \cline{3-11}
        D3 block 3 (k,L,M) & & 20, 5, 2 & 2, 1, 1 & 15, 6, 2 &  20, 5, 2 & 2, 1, 1 & 15, 6, 2  & 18, 5, 2 & 2, 1, 1 & 12, 6, 2 \\
        \hline
        down sample  & \multirow{2}{*}{$\frac{1}{8}$} &  \multicolumn{3}{c|}{avg. pool $2\times2$} & \multicolumn{3}{c|}{avg. pool $2\times2$}  & \multicolumn{3}{c}{avg. pool $2\times2$}\\
        \cline{3-11}
        D3 block 4 (k,L,M) & & 22, 5, 2 & 2, 1, 1 & 16, 7, 2 & 22, 4, 2 & 2, 1, 1 & 16, 7, 2& 20, 5, 2 & 2, 1, 1 & 14, 7, 2 \\
        \hline
        down sample  & \multirow{2}{*}{$\frac{1}{16}$} &  \multicolumn{3}{c|}{avg. pool $2\times2$} & \multicolumn{3}{c|}{avg. pool $2\times2$} &  \multicolumn{3}{c}{avg. pool $2\times2$} \\
        \cline{3-11}
        D3 block 5 (k,L,M) & & - & - & 17, 8, 2 &  - & - & 16, 8, 2 & - & - & 16, 8, 2 \\
        \hline
        up sample  & \multirow{3}{*}{$\frac{1}{8}$} & \multicolumn{3}{c|}{t.conv $2\times2$} &  \multicolumn{3}{c|}{t.conv $2\times2$} & \multicolumn{3}{c}{t.conv $2\times2$} \\
        \cline{3-11}
        concat. & & - & - & D3 block 4 & - & - & D3 block 4 & - & - & D3 block 4 \\
        D3 block 6 (k,L,M) & & -& - & 16, 6, 2& -& - & 16, 6, 2& -& - & 14, 6, 2 \\
		\hline
        up sample  & \multirow{3}{*}{$\frac{1}{4}$} & \multicolumn{3}{c|}{t.conv $2\times2$} &  \multicolumn{3}{c|}{t.conv $2\times2$} & \multicolumn{3}{c}{t.conv $2\times2$} \\
        \cline{3-11}
        concat. & & D3 block 3 & D3 block 3 & D3 block 3 & D3 block 3 & D3 block 3 & D3 block 3 & D3 block 3 & D3 block 3 & D3 block 3\\
        D3 block 7 (k,L,M) & & 20, 4, 2& 2, 1, 1 & 14, 5, 2 & 20, 4, 2& 2, 1, 1 & 14, 6, 2 & 18, 4, 2& 2, 1, 1 & 12, 6, 2\\
		\hline
        up sample  & \multirow{3}{*}{$\frac{1}{2}$} & \multicolumn{3}{c|}{t.conv $2\times2$} & \multicolumn{3}{c|}{t.conv $2\times2$}  & \multicolumn{3}{c}{t.conv $2\times2$}\\
        \cline{3-11}
        concat. & & D3 block 2 & D3 block 2 & D3 block 2 & D3 block 2 & D3 block 2 & D3 block 2 & D3 block 2 & D3 block 2 & D3 block 2 \\
        D3 block 8 (k,L,M) & & 18, 4, 2 & 2, 1, 1 & 12, 4, 2  & 18, 4, 2 & 2, 1, 1 & 12, 4, 2 & 16, 4, 2 & 2, 1, 1 & 8, 4, 2\\
		\hline
        up sample  & \multirow{3}{*}{1} & \multicolumn{3}{c|}{t.conv $2\times2$} &  \multicolumn{3}{c|}{t.conv $2\times2$}  & \multicolumn{3}{c}{t.conv $2\times2$} \\
        \cline{3-11}
        concat. & & D3 block 1 & D3 block 1 & D3 block 1 & D3 block 1 & D3 block 1 & D3 block 1 & D3 block 1 & D3 block 1 & D3 block 1\\
        D3 block 9 (k,L,M) & & 16, 4, 2 & 2, 1, 1 & 11, 4, 2  & 16, 4, 2 & 2, 1, 1 & 11, 4, 2 & 16, 4, 2 & 2, 1, 1 & 8, 4, 2\\        
        \hline
        concat. (axis) & \multirow{4}{*}{1} & \multicolumn{2}{c|}{freq} & - & \multicolumn{2}{c|}{freq} & - & \multicolumn{2}{c|}{freq} & - \\
        \cline{3-11}
        concat. (axis) &  & \multicolumn{3}{c|}{channel} &  \multicolumn{3}{c|}{channel} & \multicolumn{3}{c}{channel}\\
        \cline{3-11}
        d2 block (k,L) & & \multicolumn{3}{c|}{12, 3} & \multicolumn{3}{c|}{12, 3} & \multicolumn{3}{c}{12, 3} \\   
        \cline{3-11}
        gate conv (t$\times$f,ch) &   & \multicolumn{3}{c|}{$3\times3$, 2} &  \multicolumn{3}{c|}{$3\times3$, 2}&  \multicolumn{3}{c}{$3\times3$, 2}\\
        \hline
      \end{tabular}
    }
\end{table*}
\vspace{3mm}\\
\textbf{Results} \hspace{1mm}
The signal-to-distortion ratios (SDRs) of the proposed method and existing state-of-the-art methods are shown in Table \ref{tab:mss}. The SDRs are computed using  the {\it museval} package \cite{sisec2018} and median SDRs are reported as in the SiSEC 2018 challenge \cite{sisec2018}. 
TAK1 \cite{Takahashi18MMDenseLSTM} and UHL2 \cite{Uhlich17} are the two best performing methods in SiSEC 2018 (among submissions that do not use external data).
The proposed D3Net exhibited the best performance for \textit{vocals}, \textit{drums} and  \textit{accompaniment} (the summation of \textit{drums}, \textit{bass}, and \textit{other}) and performed comparably to the best method for \textit{other}. The average SDR of four instruments is 6.01dB, which is significantly better than all baselines. To the best of our knowledge, this is the best result reported to date.
The primaly difference between MMDenseLSTM (TAK1) and the proposed method is that MMDenseLSTM incorporates LSTM units to further expand the receptive field, whereas the proposed method uses the multidilated convolution and the nested architecture. A comparison of these methods indicates the effectiveness of the D3 block.
On the other hand, GRU dilation 1 \cite{Liu19} consists of dilated convolution and dilated GRU units without a down--up-sampling path. This also highlights the effectiveness of the dense multiresolution modeling of D3Net.
For \textit{bass}, approaches that operate in the time domain perform better, as they are capable of recovering the target phase, which is easier in the low frequency range. Among the frequency domain approaches, D3Net performs the best.

We also conduct an ablation study to validate the effectiveness of the multidilated convolution. By replacing the multidilated convolution with the standard convolution, we obtain comparable results to the best performing model in SiSEC2018, TAK1. When we replace the multidilated convolution with the standard dilated convolution, we obtain a decent improvement over D3Net without dilation even though the aliasing problem arises. However, the proposed multidilated convolution clearly outperforms the standard dilated convolution, showing the importance of handling the aliasing problem in order to incorporate dilation in DenseNet.

\section{Conclusion}
In this paper, we showed the importance of a dense multiresolution representation learning in dense prediction tasks and proposed a novel CNN architecture called D3Net. A novel multidiated convolution is introduced to enable the dense multiresolution modeling by combining with a dense skip connection topology while avoiding the aliasing problem that occurs when a standard dilated convolution is applied. We further propose a nested architecture of the densely connected multidilated convolution block to improve the parameter efficiency and provide a sufficient capacity to learn representation in each resolution. 
Extensive experiments in image semantic segmentation and audio source separation tasks confirm the effectiveness and generality of the proposed method in different types of task and domain.
D3Net outperforms state-of-the-art backbones on Cityscapes with a much smaller number of parameters. In audio source separation on MUSDB18, D3Net achieved state-of-the-art performance. 
We believe that this work provides an insight into another important property for designing CNNs: the frequency of interchanging local and global information in multiple resolutions.

{\small
\bibliographystyle{ieee_fullname}
\bibliography{audio,cv,egbib}
}
\end{document}